%% file: emnlp.tex
\documentclass[11pt,a4paper]{article}
\usepackage[hyperref]{emnlp2020}
\usepackage{times}
\usepackage{latexsym}

\usepackage{microtype}

\usepackage{url}

\usepackage{graphicx}
\usepackage{subcaption}
\usepackage{amsfonts}
\usepackage{arydshln}
\usepackage{multirow}
\usepackage{multicol}
\usepackage{amsmath} 
\renewcommand{\v}[1]{\boldsymbol{#1}} 

\usepackage{mathtools}
\usepackage[normalem]{ulem}
\usepackage{cancel}
\usepackage{booktabs}
\usepackage{comment}
\usepackage{flushend}

\usepackage[utf8]{inputenc}
\usepackage{amssymb}
\usepackage{ascii}
\usepackage{newunicodechar}
\newunicodechar{☺}{\SOH}
\newunicodechar{☻}{\STX}
\newunicodechar{♥}{\ETX}
\newunicodechar{♣}{\ENQ}
\newunicodechar{♠}{\ACK}
\newunicodechar{○}{\HT}

\usepackage{pifont}
\newcommand{\cmark}{\ding{51}}
\newcommand{\xmark}{\ding{55}}

\DeclareMathOperator*{\argmax}{argmax}

\DeclareMathOperator*{\softmax}{softmax}
\DeclareMathOperator{\relu}{ReLU}

\DeclareMathOperator{\linear}{Linear}
\DeclareMathOperator{\layernorm}{LayerNorm}
\DeclareMathOperator{\selfattn}{SelfAttn}
\DeclareMathOperator{\ffnn}{FFNN}
\DeclareMathOperator{\gru}{GRU}
\DeclareMathOperator{\cnn}{CNN}

\newcommand{\ModelName}{Ours}

\newcommand\scalemath[2]{\scalebox{#1}{\mbox{\ensuremath{\displaystyle #2}}}}

\usepackage{caption}

\usepackage{enumitem}
\aclfinalcopy 
\def\aclpaperid{3287} 

\newcommand{\squishlist}{
 \begin{list}{$\bullet$}
  { \setlength{\itemsep}{0pt}
     \setlength{\parsep}{3pt}
     \setlength{\topsep}{3pt}
     \setlength{\partopsep}{0pt}
     \setlength{\leftmargin}{1.5em}
     \setlength{\labelwidth}{1em}
     \setlength{\labelsep}{0.5em} } }

\newcounter{Lcount}
\newcommand{\squishlisttwo}{
\begin{list}{\arabic{Lcount}. }
{ \usecounter{Lcount}
\setlength{\itemsep}{0pt}
\setlength{\parsep}{0pt}
\setlength{\topsep}{0pt}
\setlength{\partopsep}{0pt}
\setlength{\leftmargin}{2em}
\setlength{\labelwidth}{1.5em}
\setlength{\labelsep}{0.5em} } }

\newcommand{\squishend}{
\end{list} }

\title{Two are Better than One:\\Joint Entity and Relation Extraction with Table-Sequence Encoders}

\author{Jue Wang$^1$ \and Wei Lu$^2$ \\
  $^1$College of Computer Science and Technology, Zhejiang University \\
  $^2$StatNLP Research Group, Singapore University of Technology and Design \\
  {\tt zjuwangjue@zju.edu.cn, luwei@sutd.edu.sg} \\}

\date{}

\begin{document}
\maketitle
\begin{abstract}
Named entity recognition and relation extraction are two important fundamental problems.
Joint learning algorithms have been proposed to solve both tasks simultaneously, {\color{black} and many of them cast the joint task as a table-filling problem.}
However, they typically focused on learning a single encoder (usually learning representation in the form of a table) to capture information required for both tasks within the same space.
We argue that it can be beneficial to design two distinct encoders to capture such two different types of information in the learning process.
In this work, we propose the novel {\em table-sequence encoders} where two different encoders -- a table encoder and a sequence encoder are designed to help each other in the representation learning process.
Our experiments confirm the advantages of having {\em two} encoders over {\em one} encoder. On several standard datasets, our model shows significant improvements over existing approaches.\footnote{
Our code is available at \url{https://github.com/LorrinWWW/two-are-better-than-one}.}

\end{abstract}

\input{1-intro}

\input{2-related}

\input{3-approach}

\input{4-exp}

\section{Conclusion}

In this paper, we introduce the novel {\em table-sequence encoders} architecture for joint extraction of entities and their relations.
It learns two separate encoders rather than one -- a sequence encoder and a table encoder where explicit interactions exist between the two encoders.
We also introduce a new method to effectively employ useful information captured by the pre-trained language models for such a joint learning task where a table representation is involved.
We achieved state-of-the-art F1 scores for both NER and RE tasks across four standard datasets, which confirm the effectiveness of our approach.
{\color{black}
In the future, we would like to investigate how the table representation may be applied to other tasks.
Another direction is to generalize the way in which the table and sequence interact to other types of representations.
}

\section*{Acknowledgements}

We would like to thank the anonymous reviewers for their helpful comments and Lidan Shou for his suggestions and support on this work. This work was done during the first author's remote internship with the StatNLP Group in Singapore University of Technology and Design. This research is supported by Ministry of Education, Singapore, under its Academic Research Fund (AcRF) Tier 2 Programme (MOE AcRF Tier 2 Award No: MOE2017-T2-1-156). 
Any opinions, findings and conclusions or recommendations expressed in this material are those of the authors and do not reflect the views of the Ministry of Education, Singapore.

\bibliography{emnlp}
\bibliographystyle{acl_natbib}


\appendix

\input{a-appendix}

\end{document}

%% file: 1-intro.tex
\section{Introduction}

Named Entity Recognition (NER, \citealt{florian2006factorizing,florian2010improving}) and
Relation Extraction (RE, \citealt{zhao2005extracting,jiang2007systematic,sun2011semi,plank2013embedding})
are two fundamental tasks in Information Extraction (IE).
Both tasks aim to extract structured information from unstructured texts.
One typical approach is to first identify entity mentions,
and next perform classification between every two mentions to extract relations,
forming a pipeline \cite{zelenko2003kernel,chan2011exploiting}.
An alternative and more recent approach is to perform these two tasks jointly \cite{li2014incremental,miwa2014modeling,miwa2016end},
which mitigates the error propagation issue associated with the pipeline approach and leverages the interaction between tasks,
resulting in improved performance.

\begin{figure}
    \centering
    \includegraphics[width=0.9\linewidth]{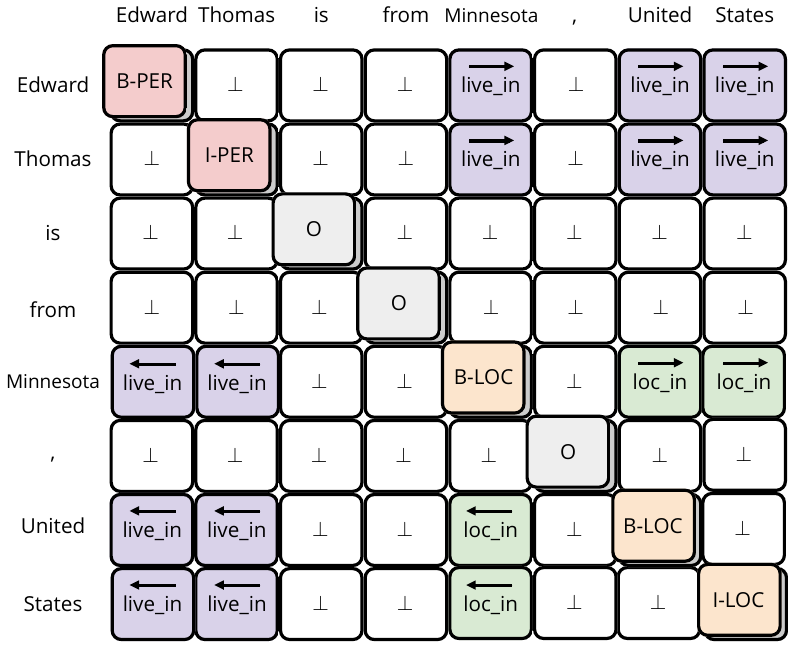}
    \caption{An example of table filling for NER and RE.}
    \label{fig:example}
\end{figure}

Among several joint approaches, one popular idea is to cast NER and RE as a table filling problem \cite{miwa2014modeling,gupta2016table,zhang2017end}.
Typically, a two-dimensional (2D) table is formed where each entry captures the interaction between two individual words within a sentence.
NER is then regarded as a sequence labeling problem where tags are assigned along the diagonal entries of the table.
RE is regarded as the problem of labeling other entries within the table.
Such an approach allows NER and RE to be performed using a single model, enabling the potentially useful interaction between these two tasks.
One example\footnote{
The exact settings for table filling may be different for different papers.
Here we fill the entire table (rather than the lower half of the table), and assign relation tags to cells involving two complete entity spans (rather than part of such spans).
We also preserve the direction of the relations.}
is illustrated in Figure \ref{fig:example}.

Unfortunately, there are limitations with the existing joint methods. First, these methods typically suffer from \emph{feature confusion} as they use a single representation for the two tasks -- NER and RE. As a result, features extracted for one task may coincide or conflict with those for the other, thus confusing the learning model.
Second, these methods \emph{underutilize} the table structure as they usually convert it to a sequence and then use a sequence labeling approach to fill the table. However, crucial structural information (e.g.,
the 4 entries at the bottom-left corner of Figure \ref{fig:example} share the same label) in the 2D table might be lost during such conversions.

In this paper, we present a novel approach to address the above limitations.
Instead of predicting entities and relations with a single representation,
we focus on learning two types of representations, namely \emph{sequence representations} and \emph{table representations}, for NER and RE respectively.
On one hand, the two separate representations can be used to capture task-specific information.
On the other hand, we design a mechanism to allow them to interact with each other, in order to take advantage of the inherent association underlying the NER and RE tasks.
In addition, we employ neural network architectures that can better capture the structural information within the 2D table representation.
As we will see, such structural information (in particular the context of neighboring entries in the table) is essential in achieving better performance.

The recent prevalence of BERT \cite{bert} has led to great performance gains on various NLP tasks.
However, we believe that the previous use of BERT, i.e., employing the contextualized word embeddings, does not fully exploit its potential.
One important observation here is that the pairwise self-attention weights maintained by BERT carry knowledge of \emph{word-word interactions}.
{Our model can effectively use such knowledge, which helps to better learn table representations.}
To the best of our knowledge, this is the first work to use the attention weights of BERT for learning table representations.

We summarize our contributions as follows:

\squishlist
\item
We propose to learn two separate encoders -- a table encoder and a sequence encoder. They interact with each other, and can capture task-specific information for the NER and RE tasks;
\item
We propose to use multidimensional recurrent neural networks to better exploit the structural information of the table representation;
\item
We effectively leverage the word-word interaction information carried in the attention weights from BERT, which further improves the performance.
\squishend

Our proposed method achieves the state-of-the-art performance on four datasets, namely ACE04, ACE05, CoNLL04, and ADE.
We also conduct further experiments to confirm the effectiveness of our proposed approach.

%% file: 2-related.tex
\section{Related Work}

NER and RE can be tackled by using separate models.
By assuming gold entity mentions are given as inputs,
RE can be regarded as a classification task.
Such models include kernel methods \cite{zelenko2003kernel}, RNNs \cite{zhang2015relation},
recursive neural networks \cite{socher2012semantic}, CNNs \cite{zeng2014relation},
and Transformer models \cite{verga2018simultaneously,wang2019extracting}.
Another branch is to detect cross-sentence level relations \cite{peng2017cross,gupta2019neural}, and even document-level relations \cite{yao2019docred,nan-etal-2020-reasoning}.
However, entities are usually not directly available in practice,
so these approaches may require an additional entity recognizer to form a pipeline.

Joint learning has been shown effective since it can alleviate the error propagation issue and benefit from exploiting the interrelation between NER and RE.
Many studies address the joint problem through a cascade approach, i.e., performing NER first followed by RE.
\citet{miwa2016end} use bi-LSTM \cite{graves2013speech} and tree-LSTM \cite{tai2015improved} for the joint task.
\citet{bekoulis2018adversarial,bekoulis2018joint} formulate it as a head selection problem.
\citet{nguyen2019end} apply biaffine attention \cite{dozat2016deep} for RE.
\citet{luan2019general}, \citet{dixit2019span}, and \citet{wadden2019entity} use span representations to predict relations.

\citet{miwa2014modeling} tackle joint NER and RE as from a table filling perspective,
where the entry at row $i$ and column $j$ of the table corresponds to the pair of $i$-th and $j$-th word of the input sentence.
The diagonal of the table is filled with the entity tags and the rest with the relation tags indicating possible relations between word pairs.
Similarly, \citet{gupta2016table} employ a bi-RNN structure to label each word pair.
\citet{zhang2017end} propose a global optimization method to fill the table.
\citet{tran2019neural} investigate CNNs on this task.

Recent work \cite{luan2019general,dixit2019span,wadden2019entity,li2019entity,eberts2019span}
usually leverages pre-trained language models such as ELMo \cite{elmo}, BERT \cite{bert}, RoBERTa \cite{roberta}, and ALBERT \cite{albert}.
However, none of them use pre-trained attention weights, which convey rich relational information between words.
We believe it can be useful for learning better table representations for RE.


%% file: 3-approach.tex
\begin{figure}[t!]
    \centering

    \includegraphics[width=\linewidth]{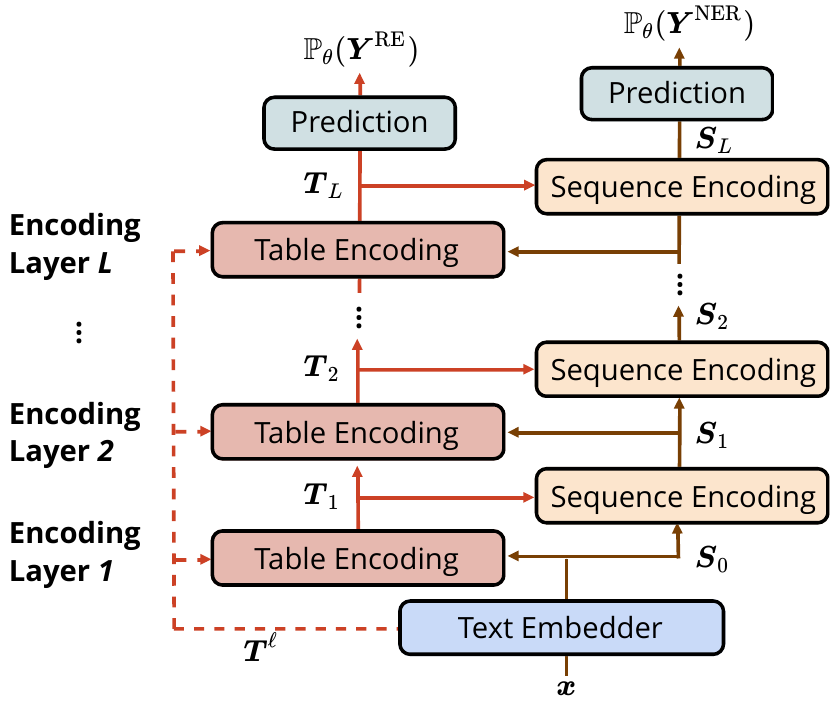}

    \caption{
        Overview of the table-sequence encoders.
        Dashed lines are for optional components ($\v T^{\ell}$).}
    \label{fig:network}

\end{figure}

\section{Problem Formulation}

In this section, we formally formulate the  NER and RE tasks.
We regard NER as a sequence labeling problem, where the gold entity tags $\v y^{\text{NER}}$ are in the standard BIO (Begin, Inside, Outside) scheme \cite{sang1999representing,ratinov2009design}.
For the RE task, we mainly follow the work of \citet{miwa2014modeling} to formulate it as a table filling problem.
Formally, given an input sentence  $\v x = [x_i]_{1 \le i \le N}$,
we maintain a tag table $\v y^{\text{RE}} = [y^{\text{RE}}_{i,j}]_{1 \le i,j \le N}$.
Suppose there is a relation with type $r$ pointing from mention $x_{i^b},..,x_{i^e}$ to mention $x_{j^b},..,x_{j^e}$,
we have $y^{\text{RE}}_{i,j} = \overrightarrow{r}$ and $y^{\text{RE}}_{j,i} = \overleftarrow{r}$ for all $i \in [i^b , i^e] \wedge j \in [j^b, j^e]$.
We use $\bot$ for word pairs with no relation.
An example was given earlier in Figure \ref{fig:example}.


\section{Model}

\begin{figure}[t!]
    \centering
    \includegraphics[width=0.8\linewidth]{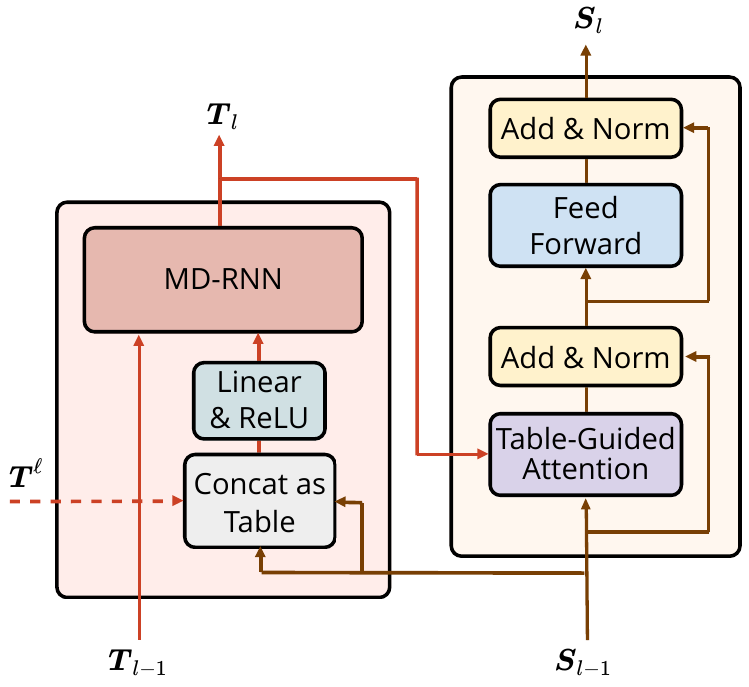}
    \caption{A layer in the table-sequence encoders.}
    \label{fig:encoder}
\end{figure}

We describe the model in this section.
The model consists of two types of interconnected encoders, a table encoder for table representation and a sequence encoder for sequence representation, as shown in Figure \ref{fig:network}.
Collectively, we call them {\em table-sequence encoders}.
Figure \ref{fig:encoder} presents the details of each layer of the two encoders, and how they interact with each other.
In each layer, the table encoder uses the sequence representation to construct the table representation;
and then the sequence encoder uses the table representation to contextualize the sequence representation.
With multiple layers, we incrementally improve the quality of both representations.


\subsection{Text Embedder}

For a sentence containing $N$ words $\v x = [x_i]_{1 \le i \le N}$, we define the word embeddings $\v x^w \in \mathbb{R}^{N \times d_1}$, as well as character embeddings $\v x^c \in \mathbb{R}^{N \times d_2}$ computed by an LSTM \cite{lample2016neural}.
We also consider the contextualized word embeddings $\v x^\ell \in \mathbb{R}^{N \times d_3}$, which can be produced from language models such as BERT.

We concatenate those embeddings for each word and use a linear projection to form the initial sequence representation $\v{S}_0 \in \mathbb{R}^{N \times H}$:
\begin{equation}
\v{S}_0= \linear([\v x^{c}; \v x^{w}; \v x^{\ell}])
\end{equation}
where each word is represented as an $H$ dimensional vector.


\subsection{Table Encoder} \label{sec:table}

The table encoder, shown in the left part of Figure \ref{fig:encoder}, is a neural network used to learn a table representation, an $N \times N$ table of vectors, where the vector at row $i$ and column $j$ corresponds to the $i$-th and $j$-th word of the input sentence.

We first construct a non-contextualized table  by concatenating every two vectors of the sequence representation followed by a fully-connected layer to halve the hidden size.
Formally, for the $l$-th layer, we have $\v{X}_l \in \mathbb{R}^{N \times N \times H}$, where:
\begin{equation}
    X_{l,i,j} = \relu(\linear([S_{l-1,i};S_{l-1,j}])) \label{math:input_mdrnn}
\end{equation}

Next, we use the Multi-Dimensional Recurrent Neural Networks (MD-RNN, \citealt{graves2007multi}) with Gated Recurrent Unit (GRU, \citealt{gru}) to contextualize $\v{X}_l$.
We iteratively compute the hidden states of each cell to form the contextualized table representation $\v{T}_{l}$, where:
\begin{equation}
    \resizebox{0.89\hsize}{!}{$T_{l,i,j} = \gru(X_{l,i,j}, T_{l-1,i,j}, T_{l,i-1,j}, T_{l,i,j-1})$} \label{math:mdrnn}
\end{equation}
We provide the multi-dimensional adaptations of GRU in Appendix \ref{sec:mdrnn} to avoid excessive formulas here.

Generally, it exploits the context along \emph{layer}, \emph{row}, and \emph{column} dimensions.
That is, it does not consider only the cells at neighbouring rows and columns,
but also those of the previous layer.

\begin{figure}
    \centering
    \includegraphics[width=\linewidth]{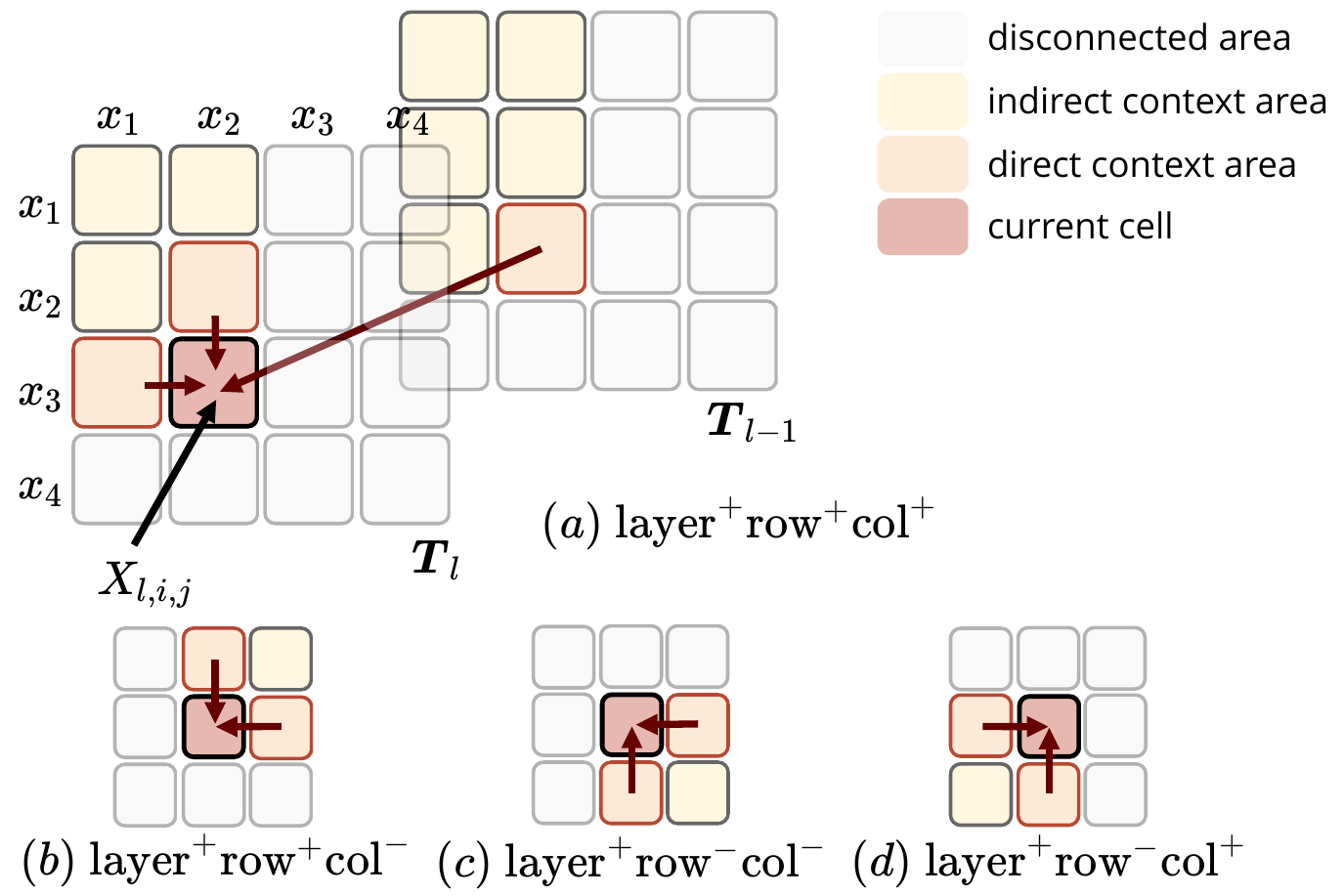}
    \caption{How the hidden states are computed in MD-RNN with 4 directions.
    We use ${D}^+$ or ${D}^-$ to indicate the direction that the hidden states flow between cells at the ${D}$ dimension (where $D$ can be \emph{layer}, \emph{row} or \emph{col}).
    For brevity, we omit the input and the \emph{layer} dimension for cases (b), (c) and (d), as they are the same as (a).}
    \label{fig:mdrnn}
\end{figure}

{The time complexity of the naive implementation (i.e., two for-loops) for each layer is $O(N \times N)$ for a sentence with length $N$.
However, antidiagonal entries\footnote{
We define antidiagonal entries to be entries at position $(i,j)$ such that $i+j=N+1+\Delta$,
where $\Delta \in [-N+1, N-1]$ is the offset to the main antidiagonal entries.
} can be calculated at the same time as they do not depend on each other.
Therefore, we can optimize it through parallelization and reduce the effective time complexity to $O(N)$.}

The above illustration describes a unidirectional RNN, corresponding to Figure \ref{fig:mdrnn}(a).
Intuitively, we would prefer the network to have access to the surrounding context in all directions.
However, this could not be done by one single RNN.
For the case of 1D sequence modeling, this problem is resolved by introducing bidirectional RNNs.
\citet{graves2007multi} discussed quaddirectional RNNs to access the context from four directions for modeling 2D data.
Therefore, similar to 2D-RNN, we also need to consider RNNs in four directions\footnote{In our scenario, there is an additional layer dimension. However, as the model always traverses from the first layer to the last layer, only one direction shall be considered for the layer dimension.}.
We visualize them in Figure \ref{fig:mdrnn}.

Empirically, we found the setting only considering cases (a) and (c) in Figure \ref{fig:mdrnn} achieves no worse performance than considering four cases altogether.
Therefore, to reduce the amount of computation, we use such a setting as default.
The final table representation is then the concatenation of the hidden states of the two RNNs:
\begin{align}
    &\resizebox{0.891\hsize}{!}{$T^{(a)}_{l,i,j} = \gru^{(a)}(X_{l,i,j}, T^{(a)}_{l-1,i,j}, T^{(a)}_{l,i-1,j}, T^{(a)}_{l,i,j-1})$} \\
    &\resizebox{0.891\hsize}{!}{$T^{(c)}_{l,i,j} = \gru^{(c)}(X_{l,i,j}, T^{(c)}_{l-1,i,j}, T^{(c)}_{l,i+1,j}, T^{(c)}_{l,i,j+1})$} \\
    &T_{l,i,j} = [T^{(a)}_{l,i,j}; T^{(c)}_{l,i,j}]
\end{align}

\subsection{Sequence Encoder}

The sequence encoder is used to learn the sequence representation -- a sequence of vectors,
where the $i$-th vector corresponds to the $i$-th word of the input sentence.
The architecture is similar to Transformer \cite{transformer}, shown in the right portion of Figure \ref{fig:encoder}.
However, we replace the scaled dot-product attention with our proposed {\em table-guided attention}.
Here, we mainly illustrate why and how the table representation can be used to compute attention weights.

\begin{figure}
  \includegraphics[width=0.95\linewidth]{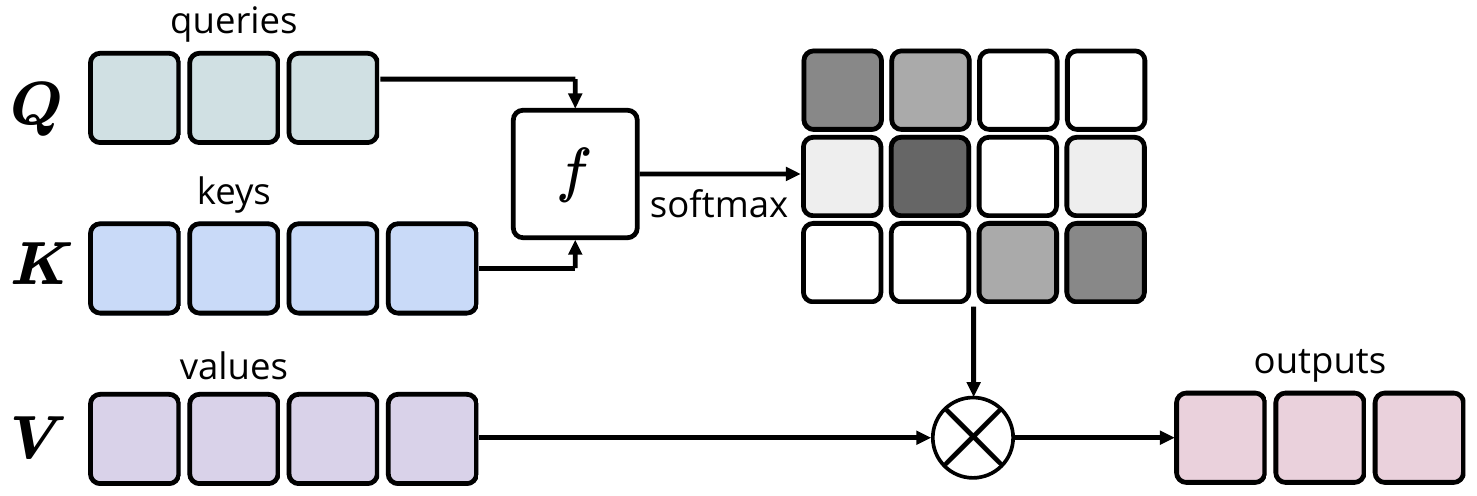}
  \caption{The generalized form of attention. The softmax function is used to normalize the weights of values $\v V$ for each query $Q_i$.}
  \label{fig:attention}
\end{figure}

First of all, given $\v Q$ (queries), $\v K$ (keys) and $\v V$ (values), a generalized form of attention is defined in Figure \ref{fig:attention}.
For each query, the output is a weighted sum of the values, where the weight assigned to each value is determined by the relevance (given by score function $f$) of the query with all the keys.

For each query $Q_i$ and key $K_j$, \citet{bahdanau2014neural} define $f$ in the form of:
\begin{align}
    f(Q_i, K_j) &= U \cdot g(Q_i, K_j) \label{math:score}
\end{align}
where $U$ is a learnable vector and $g$ is the function to map each query-key pair to a vector.
Specifically, they define $g(Q_i, K_j) = \tanh(Q_i W_0 + K_j W_1)$, where $W_0, W_1$ are learnable parameters.

Our attention mechanism is essentially a self-attention mechanism, where the queries, keys and values are exactly the same. In our case, they are essentially sequence representation $S_{l-1}$ of the previous layer (i.e., $\v Q = \v K = \v V = \v S_{l-1}$).
The attention weights (i.e., the output from the function $f$ in Figure \ref{fig:attention}) are essentially constructed from both queries and keys (which are the same in our case).
On the other hand, we also notice the table representation $\v T_l$ is also constructed from $\v S_{l-1}$. So we can consider $\v T_l$ to be a function of queries and keys, such that $T_{l,i,j} = g(S_{l-1,i}, S_{l-1,j}) = g(Q_i, K_j)$.
Then we put back this $g$ function to Equation \ref{math:score}, and get the proposed table-guided attention, whose score function is:
\begin{align}
    f(Q_i, K_j) &= U \cdot T_{l,i,j}
\end{align}

We show the advantages of using this table-guided attention:
(1) we do not have to calculate $g$ function since $\v T_l$ is already obtained from the table encoder;
(2) $\v T_l$ is contextualized along the \emph{row}, \emph{column}, and \emph{layer} dimensions, which corresponds to {queries, keys, and queries and keys in the previous layer, respectively}. Such contextual information allows the network to better capture more difficult word-word dependencies;
(3) it allows the table encoder to participate in the sequence representation learning process, thereby forming the bidirectional interaction between the two encoders.

The table-guided attention can be extended to have multiple heads \cite{transformer}, where each head is an attention with independent parameters.
We concatenate their outputs and use a fully-connected layer to get the final attention outputs.

The remaining parts are similar to Transformer.
For layer $l$, we use position-wise feedforward neural networks (FFNN) after self-attention,
and wrap attention and FFNN with a residual connection \cite{resnet} and layer normalization (\citealt{layernorm}),
to get the output sequence representation:
\begin{align}
    \tilde{\v S}_{l} &= \layernorm(\v S_{l-1} + \selfattn(\v S_{l-1})) \\
    \v S_{l} &= \layernorm(\tilde{\v S}_{l} + \ffnn(\tilde{\v S}_{l}))
\end{align}

\subsection{Exploit Pre-trained Attention Weights}

In this section, we describe the dashed lines in Figures \ref{fig:network} and \ref{fig:encoder}, which we ignored in the previous discussions.
Essentially, they exploit information in the form of attention weights from a pre-trained language model such as BERT.

We stack the attention weights of all heads and all layers to form $\v{T}^{\ell} \in \mathbb{R}^{N \times N \times (L^{\ell} \times A^{\ell})}$, where $L^{\ell}$ is the number of stacked Transformer layers, and $A^{\ell}$ is the number of heads in each layer.
We leverage $\v{T}^{\ell}$ to form the inputs of MD-RNNs in the table encoder.
Equation \ref{math:input_mdrnn} is now replaced with:
\begin{equation}
    \resizebox{0.866\hsize}{!}{$X_{l,i,j} = \relu(\linear([S_{l-1,i};S_{l-1,j};T^{\ell}_{i,j}]))$}
    \label{math:input_mdrnn2}
\end{equation}

We keep the rest unchanged.
We believe this simple yet novel use of the attention weights allows us to effectively incorporate the useful word-word interaction information captured by pre-trained models such as BERT into our table-sequence encoders for improved performance.

\section{Training and Evaluation}

We use $\v{S}_{L}$ and $\v{T}_{L}$ to predict the probability distribution of the entity and relation tags:
\begin{align}
    P_{\theta}({\v Y}^{\text{NER}})  &= \softmax(\linear(\v {S}_{L})) \\
    P_{\theta}({\v Y}^{\text{RE}} ) &= \softmax(\linear(\v {T}_{L}))
\end{align}
where ${\v Y}^{\text{NER}}$ and ${\v Y}^{\text{RE}}$ are random variables of the predicted tags,
and ${P}_{\theta}$ is the estimated probability function with $\theta$ being our model parameters.

For training, both NER and RE adopt the prevalent cross-entropy loss.
Given the input text $\v x$ and its gold tag sequence $\v y^{\text{NER}}$ and tag table $\v y^{\text{RE}}$,
we then calculate the following two losses:
\begin{align}
    \mathcal L^{\text{NER}} &= \sum_{i \in [1, N]} -\log{{P}_{\theta}(Y^{\text{NER}}_{i} = y^{\text{NER}}_{i})} \\
    \mathcal L^{\text{RE}} &= \hspace{-8px} \sum_{i,j \in [1, N]; i \ne j} \hspace{-8px} -\log{{P}_{\theta}(Y^{\text{RE}}_{i,j} = y^{\text{RE}}_{i,j})}
\end{align}

The goal is to minimize both losses $\mathcal L^{\text{NER}} + \mathcal L^{\text{RE}}$.

During evaluation, the prediction of relations relies on the prediction of entities, so we first predict the entities,
and then look up the relation probability table ${P}_{\theta}({\v Y}^{\text{RE}})$ to see if there exists a valid relation between predicted entities.

Specifically, we predict the entity tag of each word by choosing the class with the highest probability:
\begin{equation}
    \argmax_e {P}_{\theta}(Y^{\text{NER}}_{i} = e)
\end{equation}

The whole tag sequence can be transformed into entities with their boundaries and types.

Relations on entities are mapped to relation classes with highest probabilities on words of the entities.
We also consider the two directed tags for each relation.
Therefore, for two entity spans $(i^{b}, i^{e})$ and $(j^{b}, j^{e})$, their relation is given by:
\begin{align}
    \scalemath{0.89}{
    \argmax_{\overrightarrow{r}}{ 
        \hspace{-15px}
        \sum_{
            i \in [i^b , i^e], j \in [j^b , j^e]
        } 
        \hspace{-20px}
        {P}_{\theta}(Y^{\text{RE}}_{i,j} = \overrightarrow{r})
        + {P}_{\theta}(Y^{\text{RE}}_{j,i} = \overleftarrow{r} )
    }
    }
\end{align}
{where the no-relation type $\bot$ has no direction, so if $\overrightarrow{r} = \bot$, we have $\overleftarrow{r} = \bot$ as well.}

%% file: 4-exp.tex
\section{Experiments}
\subsection{Data}

We evaluate our model on four datasets, namely
ACE04 \cite{ace04}, ACE05 \cite{ace05}, CoNLL04 \cite{conll04} and ADE \cite{ade}.
More details could be found in Appendix \ref{sec:data}.

Following the established line of work, we use the F1 measure to evaluate the performance of NER and RE.
For NER, an entity prediction is correct if and only if its type and boundaries both match with those of a gold entity.\footnote{
Follow \citet{li2014incremental,miwa2016end}, we use head spans for entities in ACE. And we keep the full mention boundary for other corpora.}
For RE, a relation prediction is considered correct if its relation type and the boundaries of the two entities match with those in the gold data.
We also report the strict relation F1 (denoted RE{+}), where a relation prediction is considered correct if its relation type as well as the boundaries and types of the two entities all match with those in the gold data.
Relations are asymmetric, so the order of the two entities in a relation matters.

\subsection{Model Setup}

We tune hyperparameters based on results on the development set of ACE05 and use the same setting for other datasets.
GloVe vectors \cite{pennington2014glove} are used to initialize word embeddings.
We also use the BERT variant -- ALBERT as the default pre-trained language model.
Both pre-trained word embeddings and language model are fixed without fine-tuning.
In addition, we stack three encoding layers ($L=3$) with independent parameters including the GRU cell in each layer.
For the table encoder, we use two separate MD-RNNs with the directions of ``layer$^+$row$^+$col$^+$'' and ``layer$^+$row$^-$col$^-$'' respectively.
For the sequence encoder, we use eight attention heads to attend to different representation subspaces.
We report the averaged F1 scores of 5 runs for our models.
For each run, we keep the model that achieves the highest averaged entity F1 and relation F1 on the development set, and evaluate and report its score on the test set.
Other hyperparameters could be found in Appendix \ref{sec:training}.

\begin{table}[t!]
\centering
\tabcolsep=3px 
\newcommand{\mmicro}{${\scriptstyle\triangledown}$}
\newcommand{\mmacro}{${\scriptstyle\blacktriangle}$}
\scalebox{0.82}
{
\begin{tabular}{clccc}
\toprule
Data    & Model                 & NER & RE & RE{+}  \\ \midrule
\multirow{8}{*}{\rotatebox[origin=c]{90}{ACE04}}
    & \citet{li2014incremental}        \mmicro      & 79.7              & 48.3          & 45.3  \\ 
    & \citet{katiyar2017going}         \mmicro      & 79.6              & 49.3          & 45.7  \\ 
    & \citet{bekoulis2018joint}        \mmicro      & 81.2              & -             & 47.1  \\ 
    & \citet{bekoulis2018adversarial}  \mmicro      & 81.6              & -             & 47.5  \\ 
    & \citet{miwa2016end}              \mmicro      & 81.8              & -             & 48.4  \\ 
    & \citet{li2019entity}             \mmicro      & 83.6              & -             & 49.4  \\ 
    & \citet{luan2019general}          \mmicro      & 87.4              & 59.7          & -     \\ \cmidrule(l{5pt}r{5pt}){2-5} 
    & \ModelName{}                     \mmicro      & \textbf{88.6}     & \textbf{63.3} & \textbf{59.6}      \\ 
\midrule 
\multirow{10}{*}{\rotatebox[origin=c]{90}{ACE05}}
    & \citet{li2014incremental}        \mmicro       & 80.8              & 52.1          &  49.5   \\ 
    & \citet{miwa2016end}              \mmicro       & 83.4              & -             &  55.6   \\ 
    &  \citet{katiyar2017going}        \mmicro       & 82.6              & 55.9          &  53.6   \\ 
    &  \citet{zhang2017end}            \mmicro       & 83.6              & -             &  57.5   \\ 
    &  \citet{sun2018extracting}       \mmicro       & 83.6              & -             &  59.6   \\ 
    &  \citet{li2019entity}            \mmicro       & 84.8              & -             &  60.2   \\ 
    &  \citet{dixit2019span}           \mmicro       & 86.0              & 62.8          & -       \\ 
    &  \citet{luan2019general}         \mmicro       & 88.4              & 63.2          & -       \\ 
    &  \citet{wadden2019entity}        \mmicro       & 88.6              & 63.4          & -       \\ \cmidrule(l{5pt}r{5pt}){2-5} 
    &  \ModelName{}                    \mmicro       & \textbf{89.5}     & \textbf{67.6} & \textbf{64.3}     \\ 
\midrule
\multirow{11}{*}{
\rotatebox[origin=c]{90}{CoNLL04}
}   
    &  \citet{miwa2014modeling}\mmicro         & 80.7          & - & 61.0      \\
    &  \citet{bekoulis2018adversarial}\mmacro  & 83.6          & - & 62.0      \\
    &  \citet{bekoulis2018joint}\mmacro        & 83.9          & - & 62.0      \\
    &  \citet{tran2019neural}\mmacro           & 84.2          & - & 62.3      \\
    &  \citet{nguyen2019end}\mmacro            & 86.2	       & - & 64.4      \\
    &  \citet{zhang2017end}\mmicro             & 85.6	       & - & 67.8      \\
    &  \citet{li2019entity}\mmicro             & 87.8          & - & 68.9      \\
    &  \citet{eberts2019span}\mmicro           & 88.9          & - & 71.5      \\
    &  \citet{eberts2019span}\mmacro           & 86.3          & - & 72.9      \\ \cmidrule(l{5pt}r{5pt}){2-5}
    &  \ModelName{}\mmicro                     & \textbf{90.1} & \textbf{73.8}  & \textbf{73.6}      \\
    &  \ModelName{}\mmacro                     & \textbf{86.9} & \textbf{75.8} & \textbf{75.4}      \\ 
\midrule
\multirow{7}{*}{\rotatebox[origin=c]{90}{ADE}}
    &  \citet{li2016joint}               \mmacro      & 79.5          & - & 63.4      \\
    &  \citet{li2017neural}              \mmacro      & 84.6          & - & 71.4          \\
    &  \citet{bekoulis2018joint}         \mmacro      & 86.4          & - & 74.6      \\ 
    &  \citet{bekoulis2018adversarial}   \mmacro      & 86.7          & - & 75.5      \\ 
    &  \citet{tran2019neural}            \mmacro      & 87.1          & - & 77.3      \\
    &  \citet{eberts2019span}            \mmacro      & 89.3          & - & 79.2       \\ \cmidrule(l{5pt}r{5pt}){2-5}
    &  \ModelName{}                      \mmacro      & \textbf{89.7} & \textbf{80.1} & \textbf{80.1} \\
\bottomrule
\end{tabular}
}
\caption{Main results.
        \mmicro: micro-averaged F1; \mmacro: macro-averaged F1.
    }
\label{tab:main}
\end{table}

\subsection{Comparison with Other Models}

Table \ref{tab:main} presents the comparison of our model with previous methods on four datasets.
Our NER performance is increased by 1.2, 0.9, 1.2/0.6 and 0.4 absolute F1 points over the previous best results.
Besides, we observe even stronger performance gains in the RE task, which are 3.6, 4.2, 2.1/2.5 (RE{+}) and 0.9 (RE{+}) absolute F1 points, respectively.
This indicates the effectiveness of our model for jointly extracting entities and their relations.
Since our reported numbers are the average of 5 runs, we can consider our model to be achieving new state-of-the-art results.

\begin{table}[t]
\centering
\scalebox{0.82}
{
\begin{tabular}{lcccc}
\toprule
\multirow{2}{*}{LM}
          & \multicolumn{2}{c}{$+\v x^{\ell}$} & \multicolumn{2}{c}{$+\v{x}^{\ell}$  $+\v{T}^{\ell}$} \\
          \cmidrule(l{5pt}r{5pt}){2-3} \cmidrule(l{5pt}r{5pt}){4-5}
          &  NER & RE &  NER & RE \\ \midrule
ELMo    & 86.4 & 64.3 & -    & -                  \\
BERT    & 87.8 & 64.8 & 88.2 & 67.4                  \\
RoBERTa & 88.9 & 66.2 & 89.3 & 67.6                  \\
ALBERT  & 89.4 & 66.0 & 89.5 & 67.6                  \\
\bottomrule
\end{tabular}
}
\caption{Using different pre-trained language models on ACE05.
         $+\v x^{\ell}$ uses the contextualized word embeddings;
         $+\v T^{\ell}$ uses the attention weights.}
\label{tab:lm}
\end{table}

\subsection{Comparison of Pre-trained Models} \label{sec:pretrained}

In this section, we evaluate our method with different pre-trained language models,
including ELMo, BERT, RoBERTa and ALBERT,
with and without attention weights, to see their individual contribution to the final performance.

Table \ref{tab:lm} shows that, even using the relatively earlier contextualized embeddings without attention weights (ELMo $+\v x^{\ell}$), our system is still comparable to the state-of-the-art approach \cite{wadden2019entity},
which was based on BERT and achieved F1 scores of 88.6 and 63.4 for NER and RE respectively.
It is important to note that the model of \citet{wadden2019entity} was trained on the additional coreference annotations from OntoNotes \cite{weischedel2011ontonotes} before fine-tuning on ACE05.
Nevertheless, our system still achieves comparable results, showing the effectiveness of the table-sequence encoding architecture.

The overall results reported in Table \ref{tab:lm} confirm the importance of leveraging the attention weights, which bring improvements for both NER and RE tasks.
This allows the system using vanilla BERT to obtain results no worse than RoBERTa and ALBERT in relation extraction.

\subsection{Ablation Study}

We design several additional experiments to  understand the effectiveness of components in our system.
The experiments are conducted on ACE05.

We also compare different table filling settings, which are included in Appendix \ref{sec:form}.

\subsubsection{Bidirectional Interaction}

We first focus on the understanding of the necessity of modeling the bidirectional interaction between the two encoders.
Results are presented in Table \ref{tab:joint}.
``RE (gold)'' is presented so as to compare with settings that do not predict entities, where the gold entity spans are used in the evaluation.

\begin{table}[t]
\centering
\scalebox{0.82}
{
\begin{tabular}{lccc}
\toprule
Setting                                      & NER       & RE       &
RE (gold)
\\ \midrule
Default                                      & 89.5      & 67.6     & 70.4 \\
  \quad w/o Relation Loss                    & 89.4      & -        & -     \\
  \quad w/o Table Encoder                    & 88.4      & -        & -     \\
  \quad w/o Entity Loss                      & -         & -        & 69.8 \\
  \quad w/o Sequence Encoder                 & -         & -        & 69.2 \\
  \quad w/o Bi-Interaction                   & 88.2      & 66.3     & 69.2 \\
NER on diagonal                              & 89.4      & 67.1     & 70.2 \\
  \quad w/o Sequence Encoder                 & 88.6      & 67.0     & 70.2 \\
\bottomrule
\end{tabular}
}
\caption{
    Ablation of the two encoders on ACE05.
    Gold entity spans are given in RE (gold).
}
\label{tab:joint}
\end{table}

We first try optimizing the NER and RE objectives separately, corresponding to ``w/o Relation Loss'' and ``w/o Entity Loss''.
Compared with learning with a joint objective, the results of these two settings are slightly worse,
which indicates  that learning better representations for one task not only is helpful for the corresponding task, but also can be beneficial for the other task.

Next, we investigate the individual sequence and table encoder, corresponding to ``w/o Table Encoder'' and ``w/o Sequence Encoder''.
We also try jointly training the two encoders but cut off the interaction between them, which is ``w/o Bi-Interaction''.
Since no interaction is allowed in the above three settings, the table-guided attention is changed to conventional multi-head scaled dot-product attention,
and the table encoding layer always uses the initial sequence representation $\v S_0$ to enrich the table representation.
The results of these settings are all significantly worse than the default one, which indicates the importance of the bidirectional interaction between sequence and table representation in our table-sequence encoders.

We also experiment the use of the main diagonal entries of the table representation to tag entities, with results reported under ``NER on diagonal''.
This setup attempts to address NER and RE in the same encoding space, in line with the original intention of \citet{miwa2014modeling}.
By exploiting the interrelation between NER and RE, it achieves better performance compared with models without such information.
However, it is worse than our default setting.
We ascribe this to the potential incompatibility of the desired encoding space of entities and relations.
Finally, although it does not directly use the sequence representation, removing the sequence encoder will lead to performance drop for NER, which indicates the sequence encoder can help improve the table encoder by better capturing the structured information within the sequence.

\subsubsection{Encoding Layers}

\begin{table}[t!]
\centering
\tabcolsep=4px 
\scalebox{0.82}
{
\begin{tabular}{lcccccc}
\toprule
\multirow{2}{*}{\# Layers}
                  & \multicolumn{3}{c}{Shared}    & \multicolumn{3}{c}{Non-shared} \\
                  \cmidrule(l{5pt}r{5pt}){2-4} \cmidrule(l{5pt}r{5pt}){5-7}
                  & \# params & NER       & RE    & \# params & NER       & RE    \\ \midrule
$L=1$             &   2.2M    & 89.2      & 66.0  &   1.9M    & 89.2      & 66.0   \\
$L=2$             &   2.2M    & 89.5      & 67.0  &   3.2M    & 89.5      & 67.1   \\
$L=3$             &   2.2M    & 89.3      & 67.3  &   \underline{4.5M}    & \underline{89.5}      & \underline{67.6}   \\
$L=4$             &   2.2M    & 89.7      & 67.6  &   5.7M    & 89.6      & 67.7   \\
$L=5$             &   2.2M    & 89.6      & 67.6  &   7.0M    & 89.6      & 67.7   \\
\bottomrule
\end{tabular}
}
\caption{
    The performance on ACE05 with different number of layers.
    Pre-trained word embeddings and language models are not counted to the number of parameters. 
    The underlined ones are from our default setting.
}
\label{tab:stack}
\end{table}

Table \ref{tab:stack} shows the effect of the number of encoding layers, which is also the number of bidirectional interactions involved.
We conduct one set of experiments with shared parameters for the encoding layers and another set with independent parameters.
In general, the performance increases when we gradually enlarge the number of layers $L$.
Specifically, since the shared model does not introduce more parameters when tuning $L$,
we consider that our model benefits from the mutual interaction inside table-sequence encoders.
Typically, under the same value $L$, the non-shared model employs more parameters than the shared one to enhance its modeling capability, leading to better performance.
However, when $L>3$, there is no significant improvement by using non-shared model. We believe that increasing the number of layers may bring the risk of over-fitting, which limits the performance of the network. We choose to adopt the non-shared model with $L=3$ as our default setting.


\subsubsection{Settings of MD-RNN}

Table \ref{tab:encoder_new} presents the comparisons of using different dimensions and directions to learn the table representation, based on MD-RNN.
Among those settings,
``Unidirectional'' refers to an MD-RNN with direction ``layer$^+$row$^+$col$^+$'';
``Bidirectional'' uses two MD-RNNs with directions ``layer$^+$row$^+$col$^+$'' and ``layer$^+$row$^-$col$^-$'' respectively;
``Quaddirectional'' uses MD-RNNs in four directions, illustrated in Figure \ref{fig:mdrnn}.
Their results are improved when adding more directions, showing richer contextual information is beneficial.
Since the bidirectional model is almost as good as the quaddirectional one, we leave the former as the default setting.

\begin{table}[t]
\centering
\scalebox{0.82}
{
\begin{tabular}{lcc}
\toprule
Setting             & NER    & RE        \\ \midrule
Unidirectional
                    & 89.6   & 66.9      \\
\underline{Bidirectional}
                    & \underline{89.5}   & \underline{67.6}      \\
Quaddirectional
                    & 89.7   & 67.6      \\
Layer-wise only
                    & 89.3   & 63.9      \\
Bidirectional w/o column
                    & 89.5   & 67.2      \\
Bidirectional w/o row
                    & 89.3   & 67.4      \\
Bidirectional w/o layer
                    & 89.3   & 66.7  \\
\bottomrule
\end{tabular}
}
\caption{
    The effect of the dimensions and directions of MD-RNNs.
    Experiments are conducted on ACE05.
    The underlined ones are from our default setting.
}
\label{tab:encoder_new}
\end{table}

In addition, we are also curious about the contribution of \emph{layer}, \emph{row}, and \emph{column} dimensions for MD-RNNs.
We separately removed the \emph{layer}, \emph{row}, and \emph{column} dimension. As we can see, the results are all lower than the original model without removal of any dimension.
``Layer-wise only'' removed \emph{row} and \emph{col} dimensions, and is worse than others as it does not exploit the sentential context.

More experiments with more settings are presented in Appendix \ref{sec:context}.
Specifically, all unidirectional RNNs are consistently worse than others,
while bidirectional RNNs are usually on-par with quaddirectional RNNs.
Besides, we also tried to use CNNs to implement the table encoder.
However, since it is usually difficult for CNNs to learn long-range dependencies, we found the performance was worse than the RNN-based models.

\begin{figure*}
    \centering
    \includegraphics[width=\linewidth]{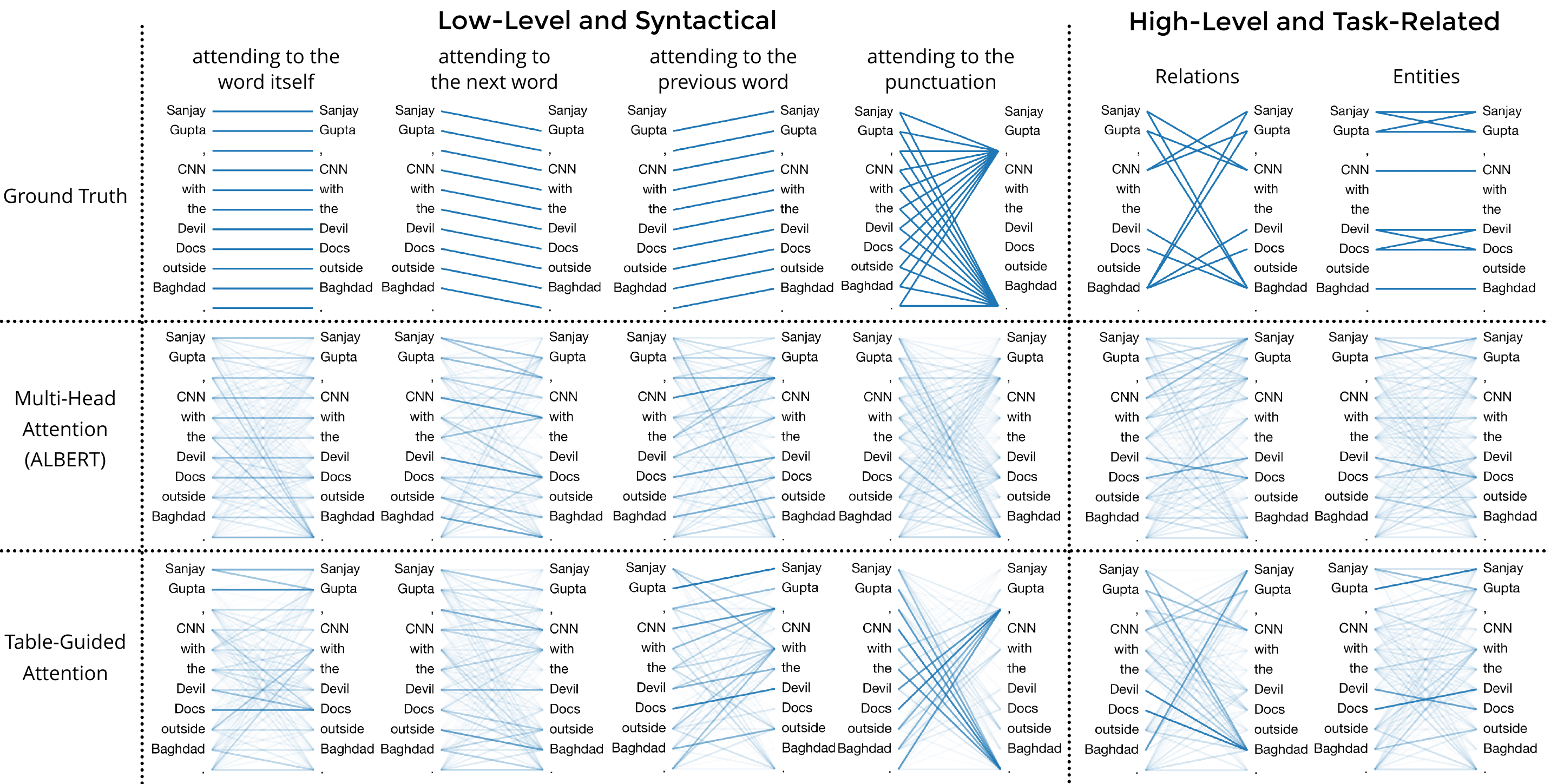}
    \caption{
        Comparison between ground truth and selected heads of ALBERT and table-guided attention.
        The sentence is randomly selected from the development set of ACE05.
    }
    \label{fig:vis}
\end{figure*}

\subsection{Attention Visualization} \label{sec:vis}

We visualize the table-guided attention with bertviz \cite{bertviz}\footnote{\url{https://github.com/jessevig/bertviz}} for a better understanding of how the network works.
We compare it with pre-trained Transformers (ALBERT) and human-defined ground truth, as presented in Figure \ref{fig:vis}.

Our discovery is similar to \citet{clark2019does}.
Most attention heads in the table-guided attention and ALBERT show simple patterns.
As shown in the left part of Figure \ref{fig:vis}, these patterns include attending to the word itself, the next word, the last word, and the punctuation.

The right part of Figure \ref{fig:vis} also shows task-related patterns, i.e., entities and relations.
For a relation, we connect words from the head entity to the tail entity;
For an entity, we connect every two words inside this entity mention.
We can find that our proposed table-guided attention has learned more task-related knowledge compared to ALBERT.
In fact, not only does it capture the entities and their relations that ALBERT failed to capture, but it also has higher confidence.
This indicates that our model has a stronger ability to capture complex patterns other than simple ones.

\subsection{Probing Intermediate States}

\begin{figure}[t!]
    \centering
    \includegraphics[width=\linewidth]{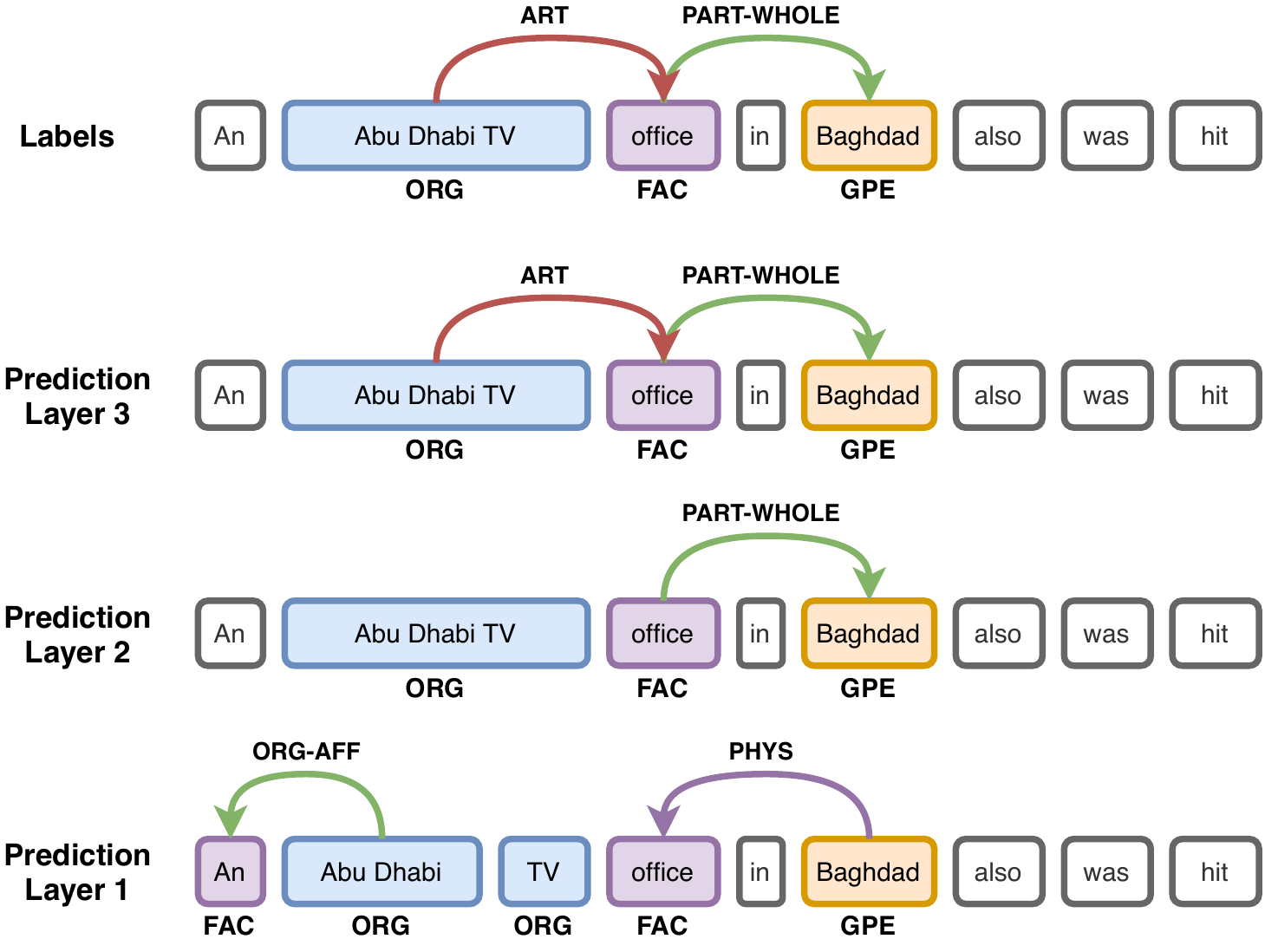}
    \caption{Probing intermediate states}
    \label{fig:main_pcase1}
\end{figure}

Figure \ref{fig:main_pcase1} presents an example picked from the development set of ACE05.
The prediction layer after training (a linear layer) is used as a probe to display the intermediate state of the model,
so we can interpret how the model improves both representations from stacking multiple layers and thus from the bidirectional interaction.
{Such probing is valid since we use skip connection between two adjacent encoding layers, so the encoding spaces of the outputs of different encoding layers are consistent and therefore compatible with the prediction layer.}

In Figure \ref{fig:main_pcase1}, the model made many wrong predictions in the first layer, which were gradually corrected in the next layers.
Therefore, we can see that more layers allow more interaction and thus make the model better at capturing entities or relations, especially difficult ones.
More cases are presented in Appendix \ref{sec:prob}.

%% file: a-appendix.tex
\section{MD-RNN} \label{sec:mdrnn}

In this section we present the detailed implementation of MD-RNN with GRU.

Formally, at the time-step layer $l$, row $i$, and column $j$, with the input $X_{l,i,j}$,
the cell at layer $l$, row $i$ and column $j$ calculates the gates as follows:
\begin{align}
    T^{prev}_{l,i,j} &= [T_{l-1,i,j}; T_{l,i-1,j}; T_{l,i,j-1}], \in \mathbb{R}^{3H} \\
    r_{l,i,j} &= \sigma([X_{l,i,j}; T^{prev}_{l,i,j}] W^r + b^r)), \in \mathbb{R}^{H} \\
    z_{l,i,j} &= \sigma([X_{l,i,j}; T^{prev}_{l,i,j}] W^z + b^z)), \in \mathbb{R}^H \\
    \tilde \lambda_{l,i,j,m} &= [X_{l,i,j}; T^{prev}_{l,i,j}] W^{\lambda}_m + b^{\lambda}_m, \in \mathbb{R}^H \\
    \lambda_{l,i,j,0}, &\lambda_{l,i,j,1}, \lambda_{l,i,j,2} = \nonumber \\
        & \softmax(\tilde \lambda_{l,i,j,0}, \tilde \lambda_{l,i,j,1}, \tilde \lambda_{l,i,j,2})
\end{align}
And then calculate the hidden states:
\begin{align}
    \tilde T_{l,i,j} &= \tanh(X_{l,i,j} {W^x}  \nonumber \\
        & + r_{l,i,j} \odot (T^{prev}_{l,i,j} {W^p}) + b^h), \in \mathbb{R}^{H}  \\
    \tilde T^{prev}_{l,i,j} &= \lambda_{l,i,j,0} \odot T_{l-1,i,j} \nonumber \\
        &+ \lambda_{l,i,j,1} \odot T_{l,i-1,j} \nonumber \\
        &+ \lambda_{l,i,j,2} \odot T_{l,i,j-1}, \in \mathbb{R}^{H} \\
    T_{l,i,j} &=
        z_{l,i,j} \odot \tilde T_{l,i,j} \nonumber \\
    & + (1-z_{l,i,j}) \odot \tilde T^{prev}_{l,i,j}, \in \mathbb{R}^{H}
\end{align}
where $W$ and $b$ are trainable parameters and
please note that they share parameters in different rows and columns but not necessarily in different layers.
Besides, $\odot$ is the element-wise product, and $\sigma$ is the sigmoid function.

As in GRU, $r$ is the \emph{reset} gate controlling whether to forget previous hidden states, and
$z$ is the \emph{update} gate, selecting whether the hidden states are to be updated with new hidden states.
In addition, we employ a \emph{lambda} gate $\lambda$, which is
used to weight the predecessor cells before passing them through the \emph{update} gate.

There are two slightly different ways to compute the candidate activation $\tilde T_{l,i,j}$, namely
\begin{align}
    \tilde T_{l,i,j} &= \tanh(X_{l,i,j} W^x \nonumber \\
    &+ r_{l,i,j} \odot (T^{prev}_{l,i,j} W^p) + b^h_l)
\end{align}
and
\begin{align}
    \tilde T_{l,i,j} &= \tanh(W^x_l X_{l,i,j} \nonumber \\
    &+ (r_{l,i,j} \odot T^{prev}_{l,i,j}) W^p + b^h_l)
\end{align}
And we found in our preliminary experiments that both of them performed as well as each other, and we choose the former, which saves some computation.

\begin{figure}[t!]
    \centering
    \includegraphics[width=0.6\linewidth]{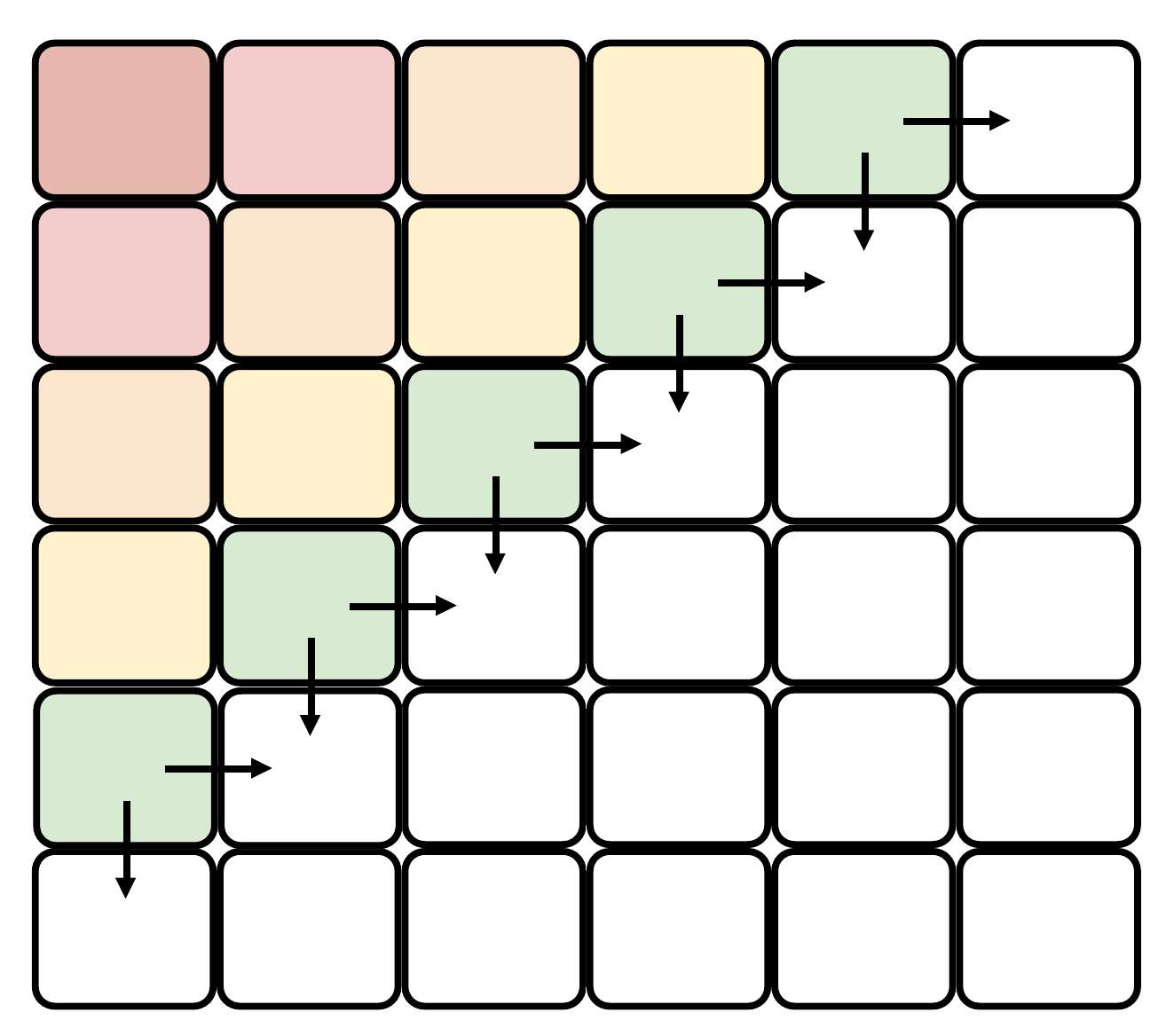}
    \caption{For 2D-RNNs, cells in the same color can be computed in parallel.}
    \label{fig:complex}
\end{figure}

The time complexity of the naive implementation (i.e., two for-loops in each layer) is $O(L\times N \times N)$ for a sentence with length $N$ and the number of encoding layer $L$.
However, antidiagonal entries can be calculated at the same time because their values do not depend on each other, shown in the same color in Figure \ref{fig:complex}.
Therefore, we can optimize it through parallelization and reduce the effective time complexity to $O(L \times N)$.

\section{Data} \label{sec:data}

\begin{table}[]
\centering
\scalebox{0.82}
{
\begin{tabular}{cccc}
    \toprule
            & \# sentences & \# entities  & \# relations     \\
            &              & (types)      & (types)         \\
    \midrule
    ACE04   & 8.7k    & 22.5k (7)    & 4.0k (6)          \\
    ACE05   & 14.5k   & 38.3k (7)    & 7.1k (6)          \\
    CoNLL04 & 1.4k    & 5.3k (4)     & 2.0k (5)          \\
    ADE     & 4.2k    & 10.5k (2)    & 6.6k (1)          \\
    \bottomrule
\end{tabular}
}
\caption{Dataset statistics}
\label{tab:data}
\end{table}

Table \ref{tab:data} shows the dataset statistics after pre-processing.
We keep the same pre-processing and evaluation standards used by most previous works.

The ACE04 and ACE05 corpora are collected from a variety of domains, such as newswire and online forums.
We use the same entity and relation types, data splits,
and pre-processing as \citet{li2014incremental} and \citet{miwa2016end}\footnote{
We use the prepocess script provided by \citet{luan2019general}:
\url{https://github.com/luanyi/DyGIE/tree/master/preprocessing}}.
Specifically, they use head spans for entities but not use the full mention boundary.

The CoNLL04 dataset provides entity and relation labels.
We use the same train-test split as \citet{gupta2016table}\footnote{
\url{https://github.com/pgcool/TF-MTRNN/tree/master/data/CoNLL04}},
and we use the same 20\% train set as development set as \citet{eberts2019span}\footnote{
\url{http://lavis.cs.hs-rm.de/storage/spert/public/datasets/conll04/}}.
Both micro and macro average F1 are used in previous work, so we will specify this while comparing with other systems.

The ADE dataset is constructed from medical reports that describe the
adverse effects arising from drug use. It contains a single relation
type ``Adverse-Effect'' and the two entity types ``Adverse-Effect'' and ``Drug''.
Similar to previous work, we filter out instances containing overlapping entities, only accounting for 2.8\% of total.

Following prior work, we perform 5-fold cross-validation for ACE04 and 10-fold for ADE.
Besides, we use 15\% of the training set as the development set.
We report the average score of 5 runs for every dataset.
For each run, we use the model that achieves the best performance (averaged entity metric score and relation metric score) on the development set, and evaluate and report its score on the test set.

\section{Hyperparameters and Pre-trained Language Models} \label{sec:training}

\begin{table}[]
    \centering
    \scalebox{0.82}
    {
        \begin{tabular}{ll}
        \toprule
        Setting                     & Value          \\
        \midrule
        batch size                  & 24             \\
        optimizer                   & Adam           \\
        learning rate (lr)          & 1e-3           \\
        warm-up steps               & 1000           \\
        dropout rate                & 0.5             \\
        \# layers (L)               & 3               \\
        \# attention heads (A)      & 8              \\
        hidden dim (H)              & 200             \\
        token emb dim               & 100            \\
        char emb dim                & 30             \\
        gradient clipping           & 5.0            \\
        \bottomrule
        \end{tabular}
    }
    \caption{Hyperparameters used in our experiments. }
    \label{tab:hyperparameters}
\end{table}

The detailed hyperparameters are present in Table \ref{tab:hyperparameters}.
For the word embeddings, we use 100-dimensional GloVe word embeddings
trained on 6B tokens\footnote{\url{https://nlp.stanford.edu/projects/glove/}} as initialization.
We disable updating the word embeddings during training.
We set the hidden size to 200, and since we use bidirectional MD-RNNs, the hidden size for each MD-RNN is 100.
We use inverse time learning rate decay: $\hat{lr} = {lr} / (1 + \text{decay\_rate} \times \text{steps} / \text{decay\_steps})$,
with decay rate 0.05 and decay steps 1000.

Besides, the tested pre-trained language models are shown as follows:
\begin{itemize}
    \item \textbf{[ELMo]} \cite{elmo}:
        Character-based pre-trained language model.
        We use the \texttt{large} checkpoint, with embeddings of dimension 3072.
    \item \textbf{[BERT]} \cite{bert}:
        Pre-trained Transformer.
        We use the \texttt{bert-large-uncased} checkpoint,
        with embeddings of dimension 1024 and attention weight feature of dimension 384 (24 layers $\times$ 16 heads).
    \item \textbf{[RoBERTa]} \cite{roberta}:
        Pre-trained Transformer.
        We use the \texttt{roberta-large} checkpoint,
        with embeddings of dimension 1024 and attention weight feature of dimension 384 (24 layers $\times$ 16 heads).
    \item \textbf{[ALBERT]} \cite{albert}:
        A lite version of BERT with shared layer parameters.
        We use the \texttt{albert-xxlarge-v1} checkpoint,
        with embeddings of dimension 4096 and attention weight feature of dimension 768 (12 layers $\times$ 64 heads).
        \textbf{We by default use this pre-trained model.}
\end{itemize}

We use the implementation provided by \citet{Wolf2019HuggingFacesTS}\footnote{\url{https://github.com/huggingface/Transformers}}
and \citet{akbik2019flair}\footnote{\url{https://github.com/flairNLP/flair}} to generate contextualized embeddings and attention weights.
Specifically, we generate the contextualized word embedding by averaging all sub-word embeddings in the last four layers;
we generate the attention weight feature (if available) by summing all sub-word attention weights for each word,
which are then concatenated for all layers and all heads.
Both of them are fixed without fine-tuning.

\section{Ways to Leverage the Table Context} \label{sec:context}

\begin{table}[]
\centering
\scalebox{0.82}
{
\begin{tabular}{lcc}
\toprule
Setting             & NER    & RE        \\ \midrule
MD-RNN && \\
\quad layer$^+$\sout{row}\ \ \sout{col}
                    & 89.3   & 63.9      \\
\quad layer$^+$row$^+$col$^+$
                    & 89.6   & 66.9      \\
\quad layer$^+$row$^+$col$^-$
                    & 89.4   & 66.3     \\
\quad layer$^+$row$^-$col$^-$
                    & 89.6   & 66.9     \\
\quad layer$^+$row$^-$col$^+$
                    & 89.4   & 66.7     \\
\quad layer$^+$row$^+$\sout{col}\ \ ; layer$^+$row$^-$\sout{col}
                    & 89.5   & 67.2      \\
\quad layer$^+$\sout{row}\ \ col$^+$; layer$^+$\sout{row}\ \ col$^-$
                    & 89.3   & 67.4      \\
\quad \sout{layer}\ \ row$^+$col$^+$; \sout{layer}\ \ row$^-$col$^-$
                    & 89.3   & 66.7  \\
\quad layer$^+$row$^+$col$^+$; layer$^+$row$^-$col$^-$
                    & 89.5   & 67.6      \\
\quad layer$^+$row$^+$col$^-$; layer$^+$row$^-$col$^+$
                    & 89.7   & 67.4      \\
\quad\begin{tabular}{@{}l@{}}
     layer$^+$row$^+$col$^+$; layer$^+$row$^-$col$^-$; \\[-5pt]
     layer$^+$row$^+$col$^-$; layer$^+$row$^-$col$^+$
\end{tabular}
                    & 89.7   & 67.6      \\
CNN && \\
\quad kernel size $1 \times 1$
                    & 89.3   & 64.7     \\
\quad kernel size $3 \times 3$
                    & 89.3   & 66.2     \\
\quad kernel size $5 \times 5$
                    & 89.3   & 65.8    \\
\bottomrule
\end{tabular}
}
\caption{Comparisons with different methods to learn the table representation.
        For MD-RNN, $D^+$, $D^-$ and $\text{\sout{$D$}}$ are indicators representing the direction, in which the hidden state flows forward, backward, or unable to flow at dimension $D$ ($D$ could be layer, row, or col).
        When using multiple MD-RNNs, we separate the indicators by ``;''.
        }
\label{tab:encoder}
\end{table}

Table \ref{tab:encoder} presents the comparisons of different ways to learn the table representation.

\vspace{4px}
\noindent
\textbf{Importance of context}
Setting ``layer$^+$\sout{row}\ \sout{col}'' does not exploit the table context when learning the table representation, instead, only layer-wise operations are used.
As a result, it performs much worse than the ones exploiting the context, confirming the importance to leverage the context information.

\vspace{4px}
\noindent
\textbf{Context along row and column}
Neighbors along both the \emph{row} and \emph{column} dimensions are important.
setting ``layer$^+$row$^+$\sout{col}\ ; layer$^+$row$^-$\sout{col}'' and ``layer$^+$\sout{row}\ col$^+$; layer$^+$\sout{row}\ col$^-$''
remove the \emph{row} and \emph{column} dimensions respectively, and their performance is though better than ``layer$^+$\sout{row}\ \sout{col}'',
but worse than setting ``layer$^+$row$^+$col$^+$; layer$^+$row$^-$col$^-$''.

\vspace{4px}
\noindent
\textbf{Multiple dimensions}
Since in setting ``layer$^+$row$^+$col$^+$'', the cell at row $i$ and column $j$ only knows the information before the $i$-th and $j$-th word,
causing worse performance than bidirectional (``layer$^+$row$^+$col$^+$; layer$^+$row$^-$col$^-$'' and ``layer$^+$row$^+$col$^-$; layer$^+$row$^-$col$^+$'') and
quaddirectional (``layer$^+$row$^+$col$^+$; layer$^+$row$^-$col$^-$; layer$^+$row$^+$col$^-$; layer$^+$row$^-$col$^+$'') settings.
Besides, the quaddirectional model does not show superior performance than bidirectional ones, so we use the latter by default.

\vspace{4px}
\noindent
\textbf{Layer dimension}
Different from the \emph{row} and \emph{column} dimensions, the \emph{layer} dimension does not carry more sentential context information.
Instead, it carries the information from previous layers, so the model can reason high-level relations based on low-level dependencies captured by predecessor layers,
which may help recognize syntactically and semantically complex relations.
Moreover, recurring along the \emph{layer} dimension can also be viewed as a layer-wise short-cut,
serving similarly to high way \cite{highway} and residual connection \cite{resnet}
and making it possible for the networks to be very deep.
By removing it (results under ``\sout{layer}\ row$^+$col$^+$; \sout{layer}\ row$^-$col$^-$''), the performance is harmed.

\vspace{4px}
\noindent
\textbf{Other network}
Our model architecture can be adapted to other table encoders.
We try CNN to encode the table representation.
For each layer $l$, given inputs $\v X_l$, we have:
\begin{align}
    \v T^0_l &= \relu(\linear([\v X_{l}; \v T_{l-1}])) \\
    \v T^1_l &= \relu(\layernorm(\cnn(\v T^0_l))) \\
    \v T_l &= \relu(\v T_{l-1} + \layernorm(\cnn(\v T^1_l)))
\end{align}

We also try different kernel sizes for CNN.
However, despite its advantages in training time, its performance is worse than the MD-RNN based ones.

\begin{table}[t!]
\centering
\scalebox{0.82}
{
\begin{tabular}{ccccc}
\toprule
\begin{tabular}{@{}c@{}}
    entire\\table?
\end{tabular}
&
\begin{tabular}{@{}c@{}}
    entire\\entity?
\end{tabular}
&
\begin{tabular}{@{}c@{}}
    directed\\relation tag?
\end{tabular}
                                            & NER       & RE       \\ \midrule
\xmark(\texttt{L})       & \xmark        & \cmark       & 89.2      & 65.9     \\
\xmark(\texttt{U})       & \xmark        & \cmark       & 89.2      & 65.8     \\
\cmark       & \xmark        & \xmark       & 89.4      & 65.1     \\
\cmark       & \cmark        & \xmark       & 89.3      & 65.8     \\
\cmark       & \xmark        & \cmark       & 89.6      & 67.1     \\
\cmark       & \cmark        & \cmark       & 89.5      & 67.6     \\
\bottomrule
\end{tabular}
}
\caption{Comparisons of different table filling formulations.
        When not filling the entire table, \texttt{L} only fills the lower-triangular part,
        and \texttt{U} fills the upper-triangular part.}
\label{tab:formulation}
\end{table}

\section{Table Filling Formulations} \label{sec:form}

Our table filling formulation does not exactly follow \citet{miwa2014modeling}.
Specifically, we fill the entire table instead of only the lower (or higger) triangular part, and
we assign relation tags to cells where entity spans intersect instead of where last words intersect.
To maintain the ratio of positive instances to negative instances, although the entire table can express directed relations by undirected tags,
we still keep the directed relation tags. I.e, if $y^{\text{RE}}_{i,j} = \overrightarrow{r}$ then $y^{\text{RE}}_{j,i} = \overleftarrow{r}$, and vice versa.
Table \ref{tab:formulation} ablates our formulation (last row), and compares it with the original one \cite{miwa2014modeling} (first row).

\begin{figure}[t!]
    \centering
    \captionsetup[subfigure]{position=b}

    \subcaptionbox{Correct the prediction at the 2nd layer
    \label{fig:pcase0}}
    {\includegraphics[height=155px]{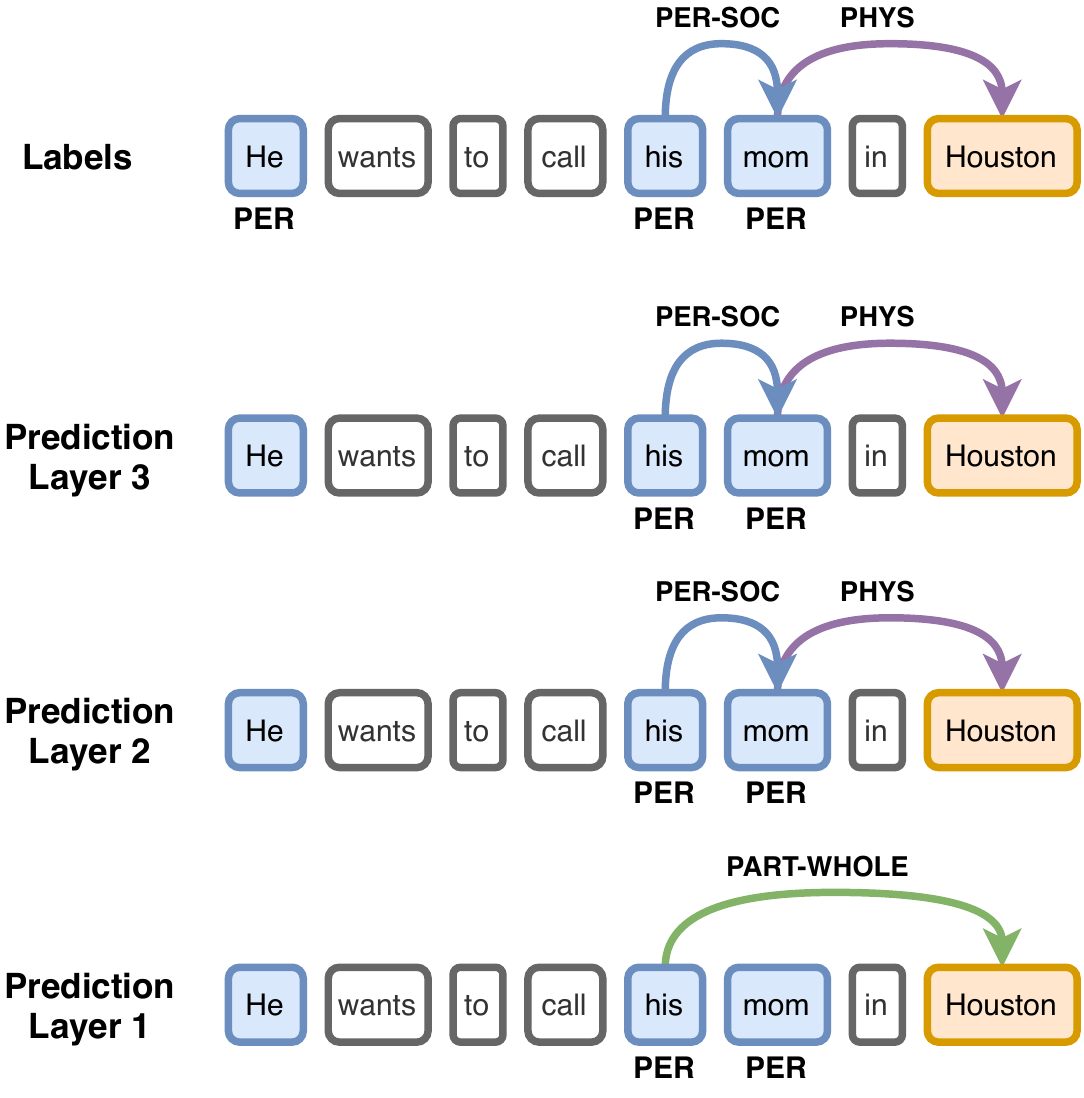}}

    \vspace{10px}
    \subcaptionbox{Correct the prediction at the 3rd layer
    \label{fig:pcase1}}
    {\includegraphics[height=155px]{pcase2_non_shared}}

    \vspace{10px}
    \subcaptionbox{A mistake at the last layer
    \label{fig:ncase0}}
    {\includegraphics[height=155px]{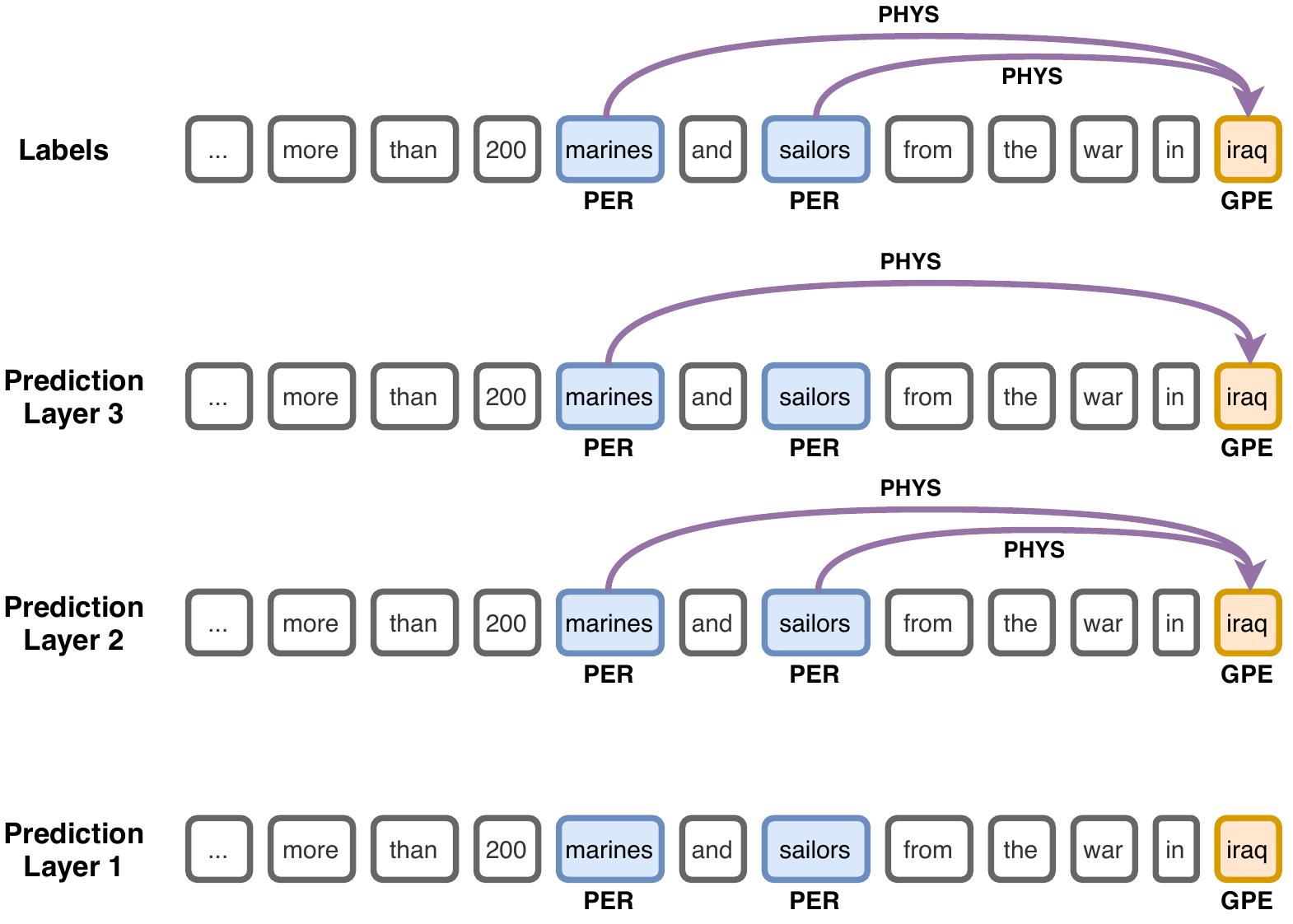}}

    \caption{Comparisons of predictions by different encoding layers.
        We predict relations and entities with the intermediate sequence and table representation,
        so that we can figure out how the model improves both representations by stacking multiple encoding layers. }
    \label{fig:cases}
    \vspace{-4mm}
\end{figure}

\section{Probing Intermediate States} \label{sec:prob}

Figure \ref{fig:cases} presents examples picked from the development set of ACE05.
The prediction layer (a linear layer) after training is used as a probe to display the intermediate state of the model,
so we can interpret how the model improves both representations from stacking multiple layers and thus from the bidirectional interaction.

Such probing is valid since
for the table encoder, the encoding spaces of different cells are consistent as they are connected through gate mechanism, including cells in different encoding layers;
for the sequence encoder, we used residual connection so the encoding spaces of the inputs and outputs are consistent.
Therefore, they are all compatible with the prediction layer.
Empirically, the intermediate layers did give valid predictions, although they are not directly trained for prediction.

In Figure \ref{fig:pcase0}, the model made a wrong prediction with the representation learned by the first encoding layer.
But after the second encoding layer, this mistake has been corrected by the model.
This is also the case that happens most frequently, indicating that two encoding layers are already good enough for most situations.
For some more complicated cases, the model needs three encoding layers to determine the final decision, shown in Figure \ref{fig:pcase1}.
Nevertheless, more layers do not always push the prediction towards the correct direction,
and Figure \ref{fig:ncase0} shows a negative example, where the model made a correct prediction in the second encoding layer, but in the end it decided not to output one relation, resulting in a false-negative error.
But we must note that such errors rarely occur, and the more common errors are that entities or relationships are not properly captured at all encoding layers.